\documentclass[runningheads]{llncs}
\usepackage[T1]{fontenc}
\usepackage{colortbl}
 \pdfoutput=1
\usepackage{array,multirow,graphicx}
\usepackage{capt-of}
\usepackage{color}
\usepackage{algorithm}
\usepackage[noend]{algpseudocode}
\usepackage{hyperref}
\usepackage{tikz}
\usepackage{comment,nicefrac}
\usepackage{graphicx}
\usepackage{capt-of}
\usepackage{bbm}
\usepackage{amsmath,amssymb}
\usepackage{booktabs}
\usepackage{epigraph,booktabs}
\usepackage{rotating}
\usepackage{tabularx}
\usepackage{adjustbox}

\usepackage{pifont}       
\usepackage{bbding}       
\usepackage{fontawesome}  

\usepackage[section]{placeins}
\usepackage{soul} 
\usepackage{color, xcolor} 

\newcommand{\ourmethod}{GEM}

\begin{document}
\title{GEM: Context-Aware Gaze EstiMation with Visual Search Behavior Matching \\ for Chest Radiograph}
\titlerunning{Context-Aware Gaze EstiMation} 

\author{Shaonan Liu$^{1}$,
Wenting Chen$^{2}$, 
Jie Liu$^{2}$, 
Xiaoling Luo$^{1,4}$ 
and Linlin Shen$^{1,3,4}$ 
}
\authorrunning{S. Liu et al}
\institute{$^1$Computer Vision Institute, College of Computer Science and Software Engineering, Shenzhen University\\
    $^2$Department of Electrical Engineering, City University of Hong Kong\\
    $^3$AI Research Center for Medical Image Analysis and Diagnosis, Shenzhen University\\
    $^4$Guangdong Provincial Key Laboratory of Intelligent Information Processing\\}
\maketitle

\begin{abstract}
Gaze estimation is pivotal in human scene comprehension tasks, particularly in medical diagnostic analysis. Eye-tracking technology facilitates the recording of physicians' ocular movements during image interpretation, thereby elucidating their visual attention patterns and information-processing strategies. In this paper, we initially define the context-aware gaze estimation problem in medical radiology report settings. To understand the attention allocation and cognitive behavior of radiologists during the medical image interpretation process, we propose a context-aware Gaze EstiMation (\ourmethod) network that utilizes eye gaze data collected from radiologists to simulate their visual search behavior patterns throughout the image interpretation process. It consists of a context-awareness module, visual behavior graph construction, and visual behavior matching. Within the context-awareness module, we achieve intricate multimodal registration by establishing connections between medical reports and images. Subsequently, for a more accurate simulation of genuine visual search behavior patterns, we introduce a visual behavior graph structure, capturing such behavior through high-order relationships (edges) between gaze points (nodes). To maintain the authenticity of visual behavior, we devise a visual behavior-matching approach, adjusting the high-order relationships between them by matching the graph constructed from real and estimated gaze points. Extensive experiments on four publicly available datasets demonstrate the superiority of GEM over existing methods and its strong generalizability, which also provides a new direction for the effective utilization of diverse modalities in medical image interpretation and enhances the interpretability of models in the field of medical imaging. \url{https://github.com/Tiger-SN/GEM}
\end{abstract}
\section{Introduction}
Eye tracking is a key technology providing gaze to understand human behavior and fundamental cognitive processes, with widespread applications in different areas~\cite{Eyewaite2019analysis,Eyevan2017visual}. Several studies have explored gaze data in various radiology fields, encompassing examinations such as Chest CT~\cite{CTaresta2020automatic}, and Knee X-rays~\cite{kneewang2022follow}. These studies provide insights into radiologists' accurate attention allocation and cognitive behavior during image interpretation, and valuable information for understanding and diagnosing. 
For instance, the distribution and density of gaze data can reveal attention patterns, allowing for precise identification of each potential lesion with a single gaze point to inform diagnostic decisions. Moreover, eye tracking can assist surgeons in dynamically adjusting their perspectives during procedures~\cite{EyeSurgenikeda2024objective}. The recordings of eye movements can also be replayed as training materials. Thus, monitoring the eye movement strategies of clinicians during image interpretation is crucial for lesion location and clinical disease diagnosis.

Automatic gaze estimation algorithms for natural scenes have been investigated in recent years. Some methods~\cite{reginschong2018connecting,reginslian2018believe} typically leverage the scene image and head image to detect gaze regions, and others~\cite{GOPwang2022gatector,HGGTRtu2022end,classtonini2023object} utilize multi-task prediction (e.g. object detection and category classification) to enhance gaze estimation accuracy. Although these methods achieve significant gaze estimation performance in natural scenes, they cannot be directly applied to clinical practice. Because radiologists often encounter difficulty in obtaining head images during medical image interpretation, models are unable to utilize crucial information regarding head tracking. In contrast, the paired reports and images are readily available, and eye gaze data can easily establish their connections. Furthermore, with the success of language models, some works have integrated pre-trained language models into medical downstream tasks~\cite{liu2023clip,yang2023tceip,chen2023fine,chen2024medical,wenting2023bi}. This inspires us to explore visual behavior patterns by establishing relationships between images and texts through eye-tracking data. Additionally, radiologists typically exhibit inherent behavioral patterns when interpreting images. For instance, they tend to first focus on prominent key lesion locations before scanning surrounding areas~\cite{Doctorhenderson2003human,Doctorbrunye2019eye}. Therefore, we have raised two main issues for the medical gaze estimation method. \textbf{Q1}: Can we establish a connection between textual content and visual perception to achieve a more accurate localization of complex and various lesions in the medical image? \textbf{Q2}: Can we design a new model architecture to simulate the visual search behavior patterns of radiologists, and thereby gain deeper insights into the visual strategies and decision-making processes of radiologists during image interpretation?

To address the aforementioned issue, we propose a \textbf{Context-Aware Gaze EstiMation(\ourmethod)} network for medical images, which first defines a medical-specific gaze estimation problem.  \ourmethod\ aims to predict the eye gaze of a radiologist with the given medical image and lesion name. \textbf{First}, to establish the connection between lesion names and medical images, we devise a \textbf{context-aware module} to achieve the fine-grained multi-modal alignment. With the fine-grained alignment, we can provide a precise textual clue for accurate gaze estimation. \textbf{Second}, to better simulate real visual search behavior patterns, we propose to use graph matching to capture and preserve the relationships among gaze points during gaze estimation. To be specific, we introduce the \textbf{visual behavior graph construction} to capture visual search behavior by representing it with the high-order relations (i.e. edge) among gaze points (i.e. nodes). To preserve the real visual behavior, we design a \textbf{visual behavior matching} to align the high-order relation between the real and estimated gaze points by matching their constructed graphs. With these devised modules, the proposed \ourmethod\ can precisely locate the specific position of the given lesion name, thus offering vital interpretation for radiologists' auxiliary analysis. Experimental results on four publicly available Chest X-ray datasets show the superiority of GEM to existing methods and exhibit impressive generalizability across both easy and hard tasks with the zero-shot setting.

\section{Method}
\subsection{Problem Definition}
Let $\{(X^i_{I}, X^i_{T}, Y^i)\}^N_{i=1}$ denotes the dataset, where $X_{T}$ represents the sentence of the medical report, $X_{I}\in\mathbb{R}^{H\times W}$ represents chest X-ray (CXR) images. $H$, and $W$ denote the height, and width of the CXR image, respectively. ${Y}\in\mathbb{R}^{K\times 2}$ stands for the coordinates $(x,y)$ of $K$ gaze points for a given sentence.  The objective of this task is to minimize the function: 
$\arg\min\limits_{\theta}\frac{1}{2N}\displaystyle\sum_{i=1}^{N}(F(X^i_I,X^i_T;\theta )-Y^i)^2$. $F$ represents the function for predicting gaze points $Y$, where $\theta$ is its corresponding parameter.

\begin{figure*}[t]
  \centering
   \includegraphics[width=\linewidth]{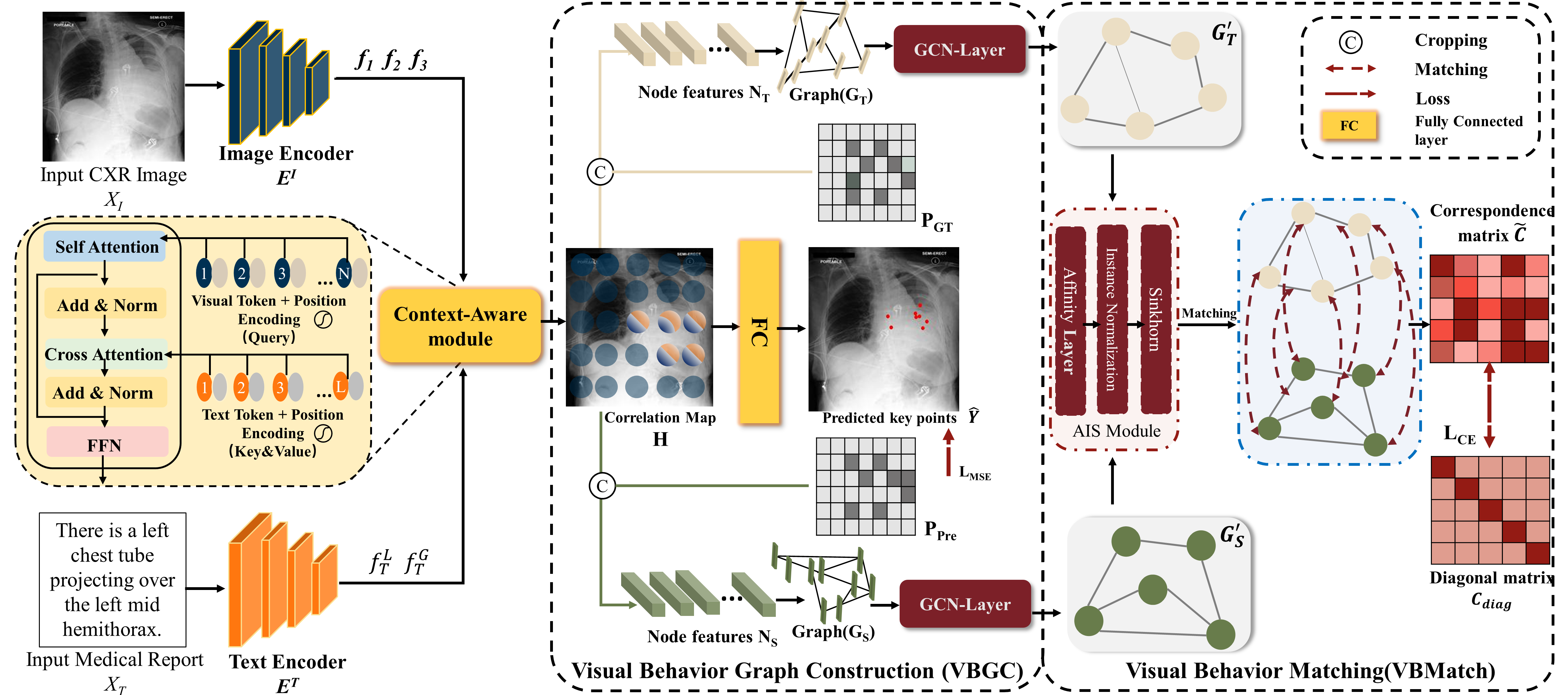}
   \caption{\textbf{Overview of the proposed Context-Aware Gaze EstiMation (GEM) network.} It consists of a context-aware module for fine-grained inter-modal alignment, a visual behavior graph construction to capture radiologists' visual search behavior, and a visual behavior matching module to preserve the behavior.}
   \label{fig:overview}
\end{figure*}

\subsection{Context-Aware Gaze Estimation}
In Fig.~\ref{fig:overview}, we propose a Context-Aware Gaze EstiMation (GEM) network to predict the eye gaze points for the medical reports. Given a CXR image ${X_{I}}$ and a sentence ${X_{T}}$ of the corresponding medical report, the image and text encoder $E^I$, $E^T$ encode them to visual and textual features, respectively. Next, the context-aware module takes these features as input and generates a correlation map $H$ to show the fine-grained multi-modal alignment relation. Then, $H$ is passed to a fully connected layer to predict gaze points $\hat{Y}$. 
In the visual behavior graph construction, to represent the behavior, we prepare the gaze masks of the ground truth and predicted gaze points $P_{GT}, P_{Pre}$, and crop the correlation map with gaze masks to obtain node features $N_T, N_S$ for their graphs $G_T, G_S$. To preserve visual behavior, we perform visual behavior matching by computing the correspondence matrix between $G_T$ and $G_S$ through AIS module and forcing it close to the diagonal matrix. $L_{MSE}$ and $L_{CE}$ are used for optimization.
\noindent\textbf{Image Encoder.} We adopt ResNet-50 as the backbone of our image encoder to obtain three separate features $f_1\in\mathbb{R}^{\frac{H}{16}\times \frac{W}{16}\times C_1}, f_2\in\mathbb{R}^{\frac{H}{8}\times \frac{W}{8}\times C_2}, f_3\in\mathbb{R}^{\frac{H}{4}\times \frac{W}{4}\times C_3}$, where $C_1, C_2, C_3$ denotes the dimension of each feature, and $H$ and $W$ denote the height and width of the CXR image, respectively.

\noindent\textbf{Text Encoder.}
We utilize the text encoder pre-trained by CLIP~\cite{CLIPradford2021learning} to extract the global textual features $f^G_{T}\in\mathbb{R}^{D}$ for the input text and the local textual features $f^L_{T}\in\mathbb{R}^{M\times D}$ for each textual token, where $D$ and $M$ denote the dimension of textual features and the number of textual tokens, respectively.

\noindent\textbf{Context-Aware Module.}
To capture the inter-modal relation in different scales, we fuse textual features with multi-scale visual features. Concretely, we first perform element-wise multiplication of $Conv(f_1)$ and $MLP(f_T^G)$ and then upscale by the factor of 2 to obtain the high-level integrated feature $\mathcal{F}_1$, where $Conv$ and $MLP$ denote the convolutional layer with kernel size of $1\times 1$ and the multilayer perceptron. For further fusion in middle scale, we integrate $f_2$ with $\mathcal{F}_1$ to obtain intermediate-level integrated features $\mathcal{F}_2=UP(Conv(f_2 \cdot \mathcal{F}_1))$, where $UP$ and $\cdot$ denote $\times 2$ upsampling layer and concatenation operation, respectively. Next, we obtain the low-level integrated features $\mathcal{F}_3=AVG(Conv(f_3 \cdot \mathcal{F}_2))$, where $AVG$ denote downsampling layer. With multi-level integrated features, we aggregate them through CoordConv layer~\cite{coordliu2018intriguing} to generate the multi-modal features $\mathcal{F}_m$. Finally, we pass $\mathcal{F}_m$ and local textual features $f_T^L$ to self-attention layer $SA$, cross-attention layer $CA$ and feed-forward network $FFN$ to obtain the correlation map $H=FFN(CA(SA(\mathcal{F}_m+PE, f_T^L+PE)))$, where $PE$ indicates the position embeddings. The correlation map shows the fine-grained alignment relation between the input CXR image and texts. 

\subsection{Visual Behavior Graph Construction (VBGC)}
As radiologists' gaze points scatter similar to a star pattern with a central point and surrounding ones, we introduce visual behavior graph construction to capture their visual behavior, which utilizes a graph to represent the high-order relations~\cite{liu2021graph} (i.e., edges) among gaze points (i.e., nodes).
Specifically, we prepare the gaze masks for the ground-truth (GT) and estimated gaze points $P_{GT}, P_{Pre}$. The gaze mask is obtained by masking $6 \times 6$ patches centered around the gaze points. To extract the gaze point features, we crop the correlation map with the gaze masks to obtain the node features for the GT and estimated gaze points $N_T, N_S$. With the node features, we obtain their graph edges $E_T$ and $E_S$ by applying an edge generator~\cite{AISfu2021robust} to the node features. The edge generator initially employs a transformer model to learn the soft edge relationships between any pair of nodes in the graph. Then, the softmax function is applied to the inner product of soft edge features to derive the soft edge adjacency matrices $E_T$ and $E_S$. These graph edges unveil the higher-order relationships between the GT and estimated gaze point features, facilitating a more profound comprehension of the interaction among features across various levels. After that, we obtain the graph structures $G_{\text{T}}=\left\{N_T; E_T\right\}$ and $G_{\text{S}}=\left\{N_S; E_S\right\}$ for GT and estimated gaze points, respectively. To capture higher-order relationships, we utilize graph convolutional networks (GCN) to embed graph nodes and higher-order graph structures (edges) into the node feature space, thereby generating new node features $GCN(N_T, E_T)$ and $GCN(N_S, E_S)$. 

\subsection{Visual Behavior Matching (VBMatch)}
To learn the actual visual behavior of radiologists, we propose a visual behavior matching to align the high-order relation between the GT and estimated gaze points by matching their graphs. To be specific, given the graphs for GT and estimated gaze points $G_T, G_S$, we perform graph matching between them to reduce the disparity in their corresponding relationships within the graph feature space. The AIS module~\cite{AISfu2021robust} is employed to calculate the soft correspondence matrix,
\begin{equation}
    \tilde{C} = \text{AIS}(GCN(N_T, E_T),GCN(N_S, E_S)),
\end{equation}
where $\tilde{C}$ represents the soft correspondence between nodes in two predicted graphs, indicating the likelihood of establishing matching relationships between any pair of nodes in the two graphs. The AIS module~\cite{AISfu2021robust} consists of an affinity layer to compute an affinity matrix between two graphs, instance normalization~\cite{insulyanov2016instance} to ensure the elements of the affinity matrix are positive, and Sinkhorn~\cite{sinkhorn1964relationship} to address outliers in the affinity matrix. To encourage the high-order relation of GT gaze points close to that of estimated ones, we force one-to-one correspondence between their nodes and edges by setting the target relation matrix as a diagonal matrix. To achieve this, we compute the cross-entropy loss $L_{CE}$ between the soft correspondence matrix $\tilde{C}$ and the target correspondence matrix $C_{diag}$. In order to provide more supervision to the estimated gaze points, we employ the Mean Squared Error (MSE) loss $L_{MSE}$ to quantify the disparity between them. The overall objective function is defined as:
\begin{equation}
    L_{\text{total}} = \alpha L_{MSE}(\hat{Y}, Y) +\beta {L_{CE}(\tilde{C},C_{diag})}.
\end{equation}

\section{Experminents and Results}
\subsection{Experiment Setting}
\noindent\textbf{Datasets.} 
We conduct the experiments on four publicly available chest X-ray (CXR) datasets, including \textbf{MIMIC-Eye}~\cite{MIMIC_Eyehsieh2023mimic}, \textbf{OpenI}~\cite{OpenIdemner2016preparing}, \textbf{MS-CXR}~\cite{MCCXRboecking2022making} and \textbf{AIforCovid}~\cite{italiansoda2021aiforcovid}.
The \textbf{MIMIC-Eye} dataset includes 8,164 pairs of CXR images, diagnostic reports, and eye-tracking data collected from radiologists while examining diagnostic reports and images from the cohort of 3,192 patients. 6,600 pairs are used for training, 800 for validation, and 764 for testing.
The \textbf{OpenI} dataset comprises 3,684 pairs of radiology reports and CXR images, while the \textbf{AIforCovid} dataset consists of the clinical data and CXR images from 820 registered patients. The \textbf{MS-CXR} dataset contains 1,047 CXR images with annotations of 1,153 bounding boxes and the corresponding sentences. The CXR images are resized to $224\times 224$.

\noindent\textbf{Implementation Details.} 
We adopt the image and text encoder pre-trained by CLIP and freeze the parameters during training.
AdamW optimizer is used with a learning rate of 1e-6 and a batch size of 16. The training epoch is 12. $\alpha$ and $\beta$ are set to 1 and 0.1. All experiments are conducted with the PyTorch on one Nvidia V100 32GB GPU.

\noindent\textbf{Evaluation Metrics.}
We utilize Mean Squared Error (MSE), Mean Absolute Error (MAE) distances, and the Probability of Correct Keypoint (PCK)~\cite{PCKyang2012articulated} to assess gaze estimation accuracy. PCK incorporates predefined thresholds of 0.2, 0.3, and 0.4 for a thorough and balanced evaluation.

\begin{figure*}[t]
  \centering
   \includegraphics[width=\linewidth]{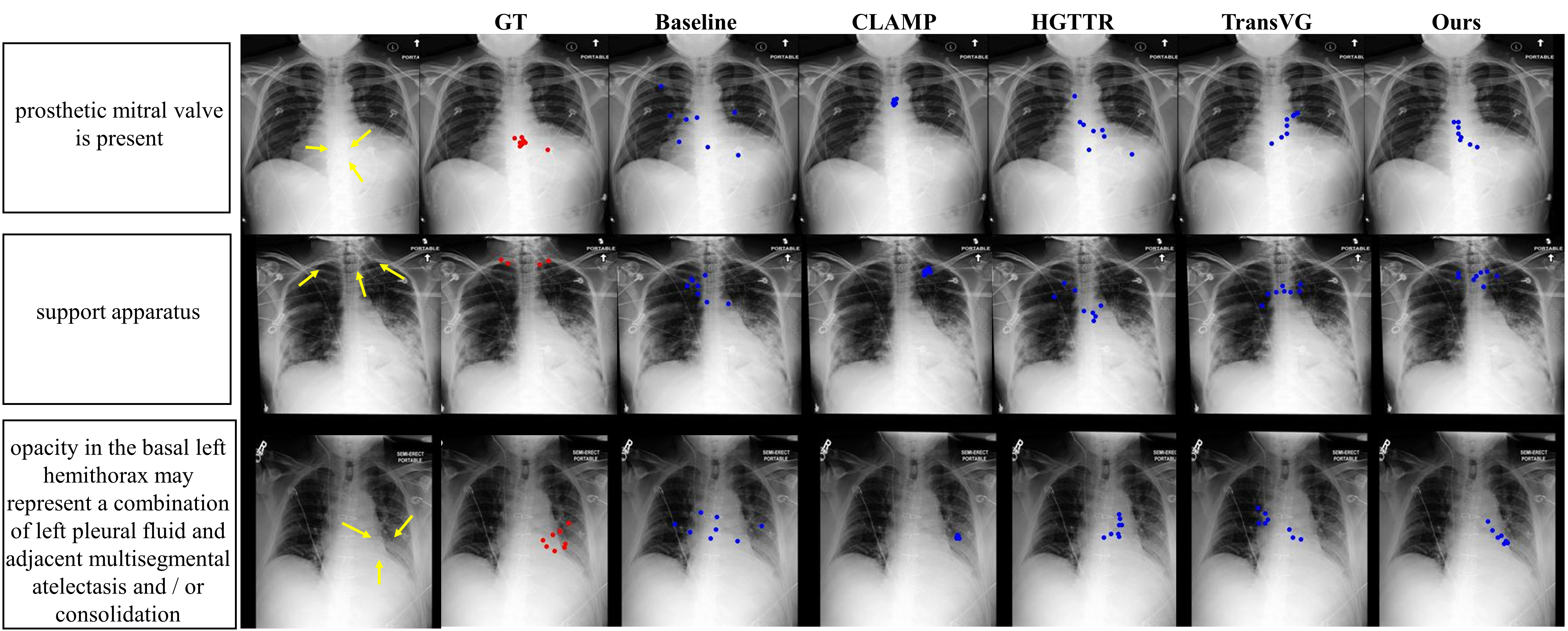}
   \caption{Qualitative comparison of \ourmethod\ and other models on MIMIC-Eye dataset. Yellow arrows indicate regions relevant to input texts, and red and blue points represent the GT and estimated gaze points, respectively.}
  \label{fig:MIMIC-Eye}
\end{figure*}
\begin{table}[t]
\small
\setlength\tabcolsep{5pt}
\renewcommand{\arraystretch}{1.1}
	\centering
	\caption{Gaze estimation performance of different methods. Asterisks indicate statistical significance: * p < 0.05, ** p < 0.01, *** p < 0.001}
\scalebox{0.94}{
 \begin{tabular}{c|ccccc}

			\toprule[1.5pt] 
    Methods&  MSE\quad & MAE\quad & PCK@0.2 & PCK@0.3 & PCK@0.4\\ \hline
Baseline & 0.0387*** & 0.156*** & 41.950*** & 68.701*** & 87.058*** \\
CLAMP~\cite{clampzhang2023clamp} & 0.0580*** & 0.182*** & 37.549*** & 57.806*** & 74.035*** \\ 
HGTTR~\cite{HGGTRtu2022end} & 0.0357*** & 0.150*** & 43.751*** & 71.174***  & 89.528**\quad\\ 
TransVG~\cite{transvgdeng2021transvg} & 0.0327*\quad \quad & 0.141*\quad \quad & 50.032*\quad \quad & 74.951** \quad   & 89.217**\quad \\ 
						
\hline
\textbf{\ourmethod} & \textbf{0.0320 \quad} & \textbf{0.139\quad} & \textbf{50.490\quad} & \textbf{77.012\quad} & \textbf{90.477\quad}\\ 
			\bottomrule[1pt]
	\end{tabular}
	}
	\label{tab:Compare}
\end{table}

\subsection{Comparison with Baseline Models}
We quantitatively and qualitatively compare our performance with the baseline model and keypoints estimation methods for natural scenes, including the CLAMP~\cite{clampzhang2023clamp} that employs text cues to estimate animal poses, TransVG~\cite{transvgdeng2021transvg} whose output layer is modified for gaze estimation, and HGGTR~\cite{HGGTRtu2022end} with text input. In Table~\ref{tab:Compare}, our GEM surpasses current methods by a large margin with the PCK@0.3 scores of 19.206 and 2.061, exhibiting the superiority of the proposed method. Fig.~\ref{fig:MIMIC-Eye} visualizes the gaze points estimated by the proposed method and other methods. The baseline model produces scattered predictions, CLAMP aggregates gaze into a single position, but TransVG's and HGGTR's estimations lack accuracy. Conversely, GEM not only predicts precise gaze points but also captures radiologists' visual search behavior, demonstrating our superiority.

\subsection{Ablation Study}
\begin{figure*}[t]
  \centering
\includegraphics[width=0.85\linewidth]{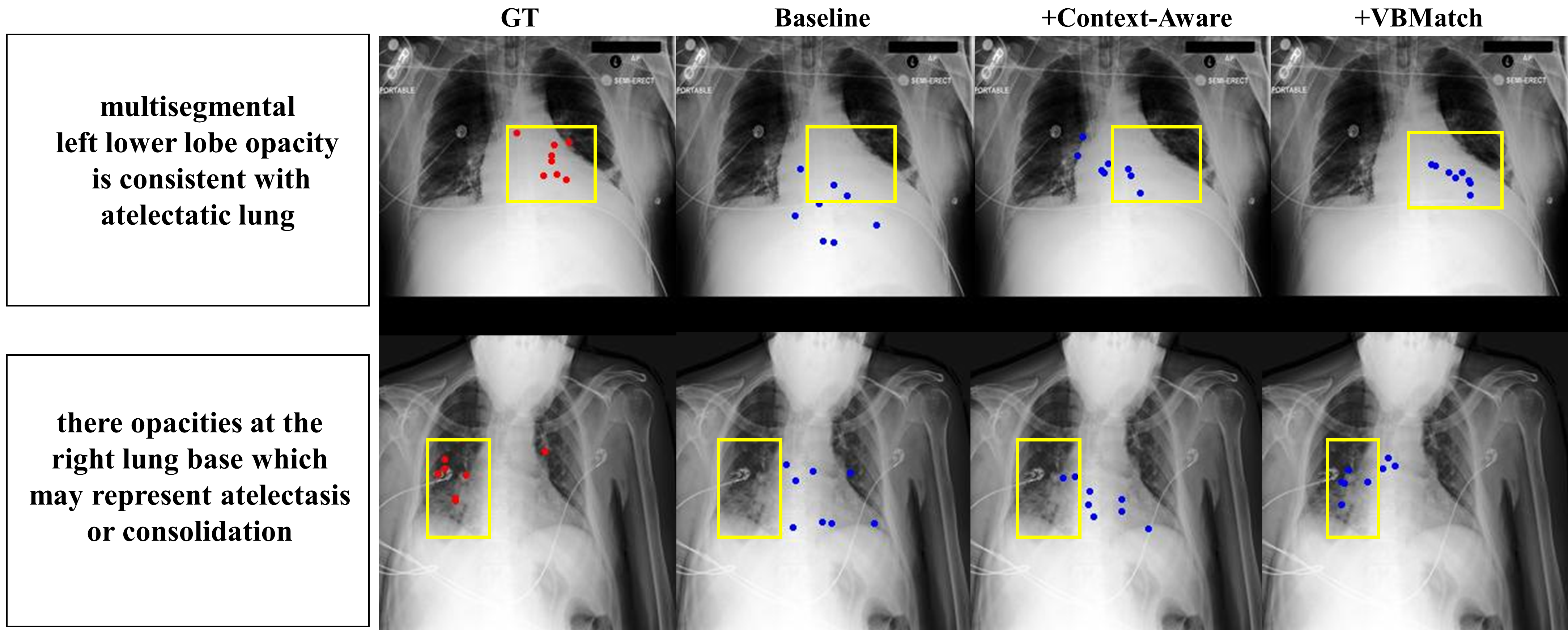}
   \caption{Visualization of the ablation study on the MIMIC-Eye dataset.}
   \label{fig:ablation}
\end{figure*}

\noindent\textbf{Effectiveness of Context-Aware Modules.}
In Table~\ref{tab:ablation}, the baseline model integrates textual to visual features through element-wise addition for gaze estimation. When using context-aware module, the performance is significantly improved by 10.166 in the PCK@0.3 score.  Fig.~\ref{fig:ablation} shows that the gaze points (blue) predicted by the baseline with context-aware module are closer to the GT ones (red), indicating the effectiveness of the context-aware module.
\begin{table}[t]
\small
\setlength\tabcolsep{2pt}
\renewcommand{\arraystretch}{1}
	\centering
	\caption{Ablation study on the MIMIC-Eye dataset.}
 \scalebox{0.82}{
 \begin{tabular}{c|cc|cccccc}

			\toprule[1pt] 
    Baseline & \multicolumn{2}{c|}{\textbf{GEM}} & \multicolumn{5}{c}{Metrics}\\ \hline 
    Addition & Context-Aware &  VBMatch  & MSE & MAE & PCK@0.2 &  PCK@0.3 & PCK@0.4 \\ \hline
\ding{52} & \ding{55} & \ding{55} & 0.0387 & 0.156 & 41.950 & 68.701 & 87.058 \\
\ding{55} & \ding{52} & \ding{55}  & 0.0322 & 0.142 & 48.020 & 75.867 & 87.009  \\ 
\ding{55} & \ding{52} & \ding{52}  & \textbf{0.0320} & \textbf{0.139} & \textbf{50.490} & \textbf{77.012} & \textbf{90.477}\\
			\bottomrule[1pt]
	\end{tabular}
	}
	\label{tab:ablation}
\end{table}

\noindent\textbf{Effectiveness of Visual Behavior Matching (VBMatch).}
After integrating both visual behavior matching (VBMatch) and context-aware module, the PCK@0.3 score remarkably increases by 1.145, as listed in Table~\ref{tab:ablation}. Fig.~\ref{fig:ablation} illustrates that further equipping the baseline model with VBMatch makes the estimated points more accurate and similar to actual search behavior, suggesting the effectiveness of VBMatch.

\subsection{Generalizability Analysis}
\begin{figure*}[t]
  \centering
   \includegraphics[width=0.82\linewidth]
   {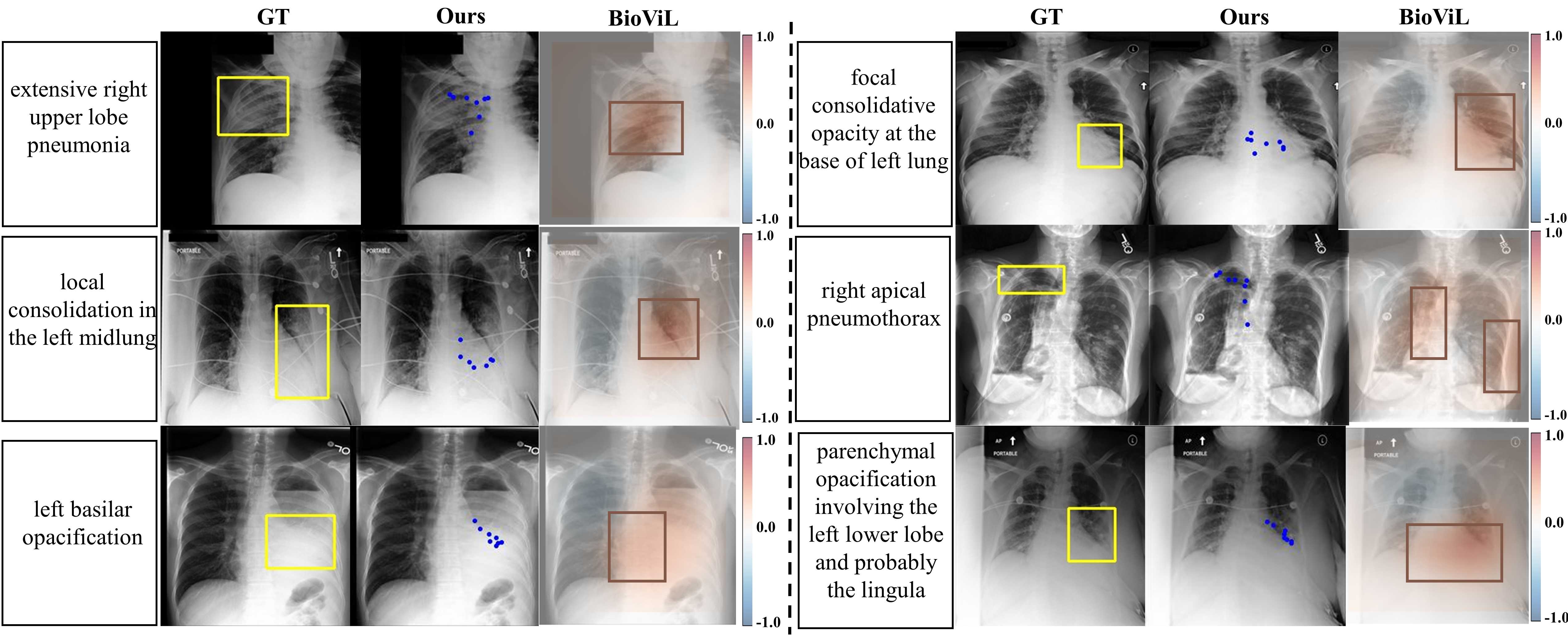}
   \caption{Qualitative evaluation of our method on the easy task of phrase grounding on the MS-CXR dataset. Yellow boxes indicate radiologists' annotations, blue points are estimated points, and the heatmap and box show focal areas highlighted by BioViL~\cite{Biovilboecking2022making}.}
  \label{fig:MS-CXR}
\end{figure*}

\begin{figure*}[t]
  \centering
   \includegraphics[width=0.8\linewidth]{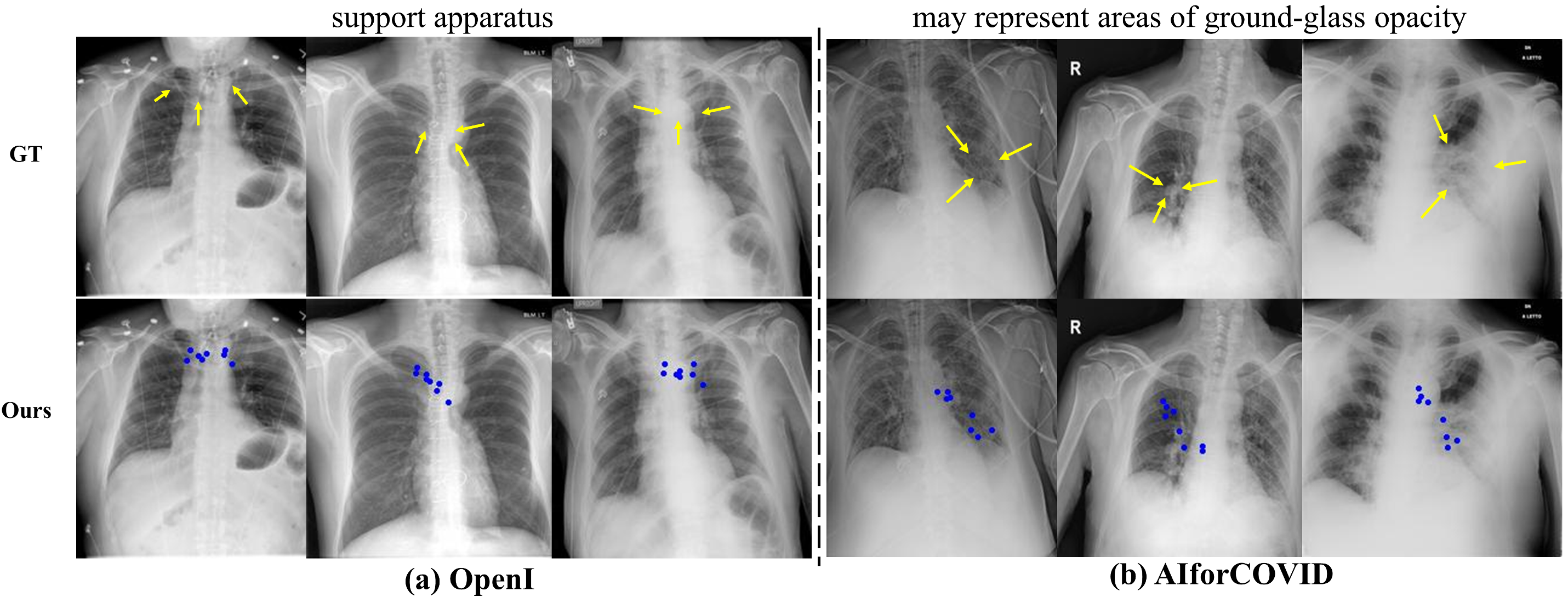} 
   \caption{Qualitative evaluation of our method on the hard task of gaze estimation on OpenI and AIforCOVID datasets. Yellow arrows indicate radiologists' annotation, and blue points represent estimated gaze points.}
   \label{fig:OpenI}
\end{figure*}

To further analyze the generalizability of our method, we design both easy and hard tasks with the zero-shot setting. The former is the phrase grounding task that predicts the rough region (i.e. bounding box) with given texts on the MS-CXR dataset~\cite{MCCXRboecking2022making}, and the latter is to estimate precise gaze points on the OpenI~\cite{OpenIdemner2016preparing} and AIforCovid~\cite{italiansoda2021aiforcovid} datasets. In the easy task, Fig.~\ref{fig:MS-CXR} shows that the gaze points generated by our method are accurately located in the GT box and comparable to detection results of BioViL~\cite{Biovilboecking2022making}, indicating that our method can generalize well on the phrase grounding task. For the hard task,
in Fig.~\ref{fig:OpenI}, our method can generate precise gaze points that highly correspond to the real annotations on both unseen datasets, implying our strong generalizability in zero-shot settings.

\section{Conclusion}
In this work, we propose a GEM network, which presents the first work to define and address the context-aware gaze estimation problem in medical scenarios. We devise a context-aware module to establish the fine-grained relation between medical images and reports to facilitate gaze estimation. To learn the radiologists' visual search behavior patterns, we propose visual behavior graph construction to represent the visual behavior with graphs and employ visual behavior matching to preserve the behavior. Extensive experiments prove the superiority of the proposed method over other models and show its strong generalizability and interpretability across easy and hard tasks with zero-shot settings.

\subsubsection{\ackname} This work was supported by the National Natural Science Foundation of China under Grant 82261138629 and 12326610; Guangdong Basic and Applied Basic Research Foundation under Grant 2023A1515010688; Guangdong Provincial Key Laboratory under Grant 2023B1212060076; Shenzhen Municipal Science and Technology Innovation Council under Grant JCYJ20220531101412030.
\subsubsection{Disclosure of Interests.} The authors have no competing interests to declare that are relevant to the content of this article.
%
%
%
\bibliographystyle{splncs04}
\bibliography{mybibliography}
%


\end{document}